\documentclass[letterpaper, 10 pt, conference]{ieeeconf}  %

\IEEEoverridecommandlockouts                              %

\usepackage{cite}
\usepackage{amsmath,amssymb,amsfonts}
\usepackage{algorithmic}
\usepackage{graphicx}
\usepackage{textcomp}
\usepackage{xcolor}

\usepackage{array, booktabs}       %
\usepackage{amsfonts}       %
\usepackage{nicefrac}       %
\usepackage{microtype}      %
\usepackage{xcolor}         %
\usepackage{colortbl}         %
\usepackage{pifont}%
\newcommand{\cmark}{\ding{51}}%
\newcommand{\xmark}{\ding{55}}%
\usepackage{mathtools}
\usepackage{math_commands}
\usepackage{lib}
\usepackage{alias}
\usepackage{wrapfig}
\usepackage{hyperref}

\title{\LARGE \bf
One-shot Visual Imitation via Attributed Waypoints \\and Demonstration Augmentation
}

\author{Matthew Chang$^1$ and Saurabh Gupta$^1$%
\thanks{$^1$University of Illinois Urbana-Champaign, Illinois, USA. Emails: \texttt{\{mc48,saurabhg\}@illinois.edu}.}
}

\begin{document}
\pagenumbering{arabic}
\pagestyle{plain}
\maketitle

\begin{abstract}
In this paper, we analyze the behavior of existing techniques and design new
solutions for the problem of one-shot visual imitation. In this setting, an agent must
solve a novel instance of a novel task given just a single visual demonstration. Our analysis reveals that current methods fall short because of three errors:
the DAgger problem arising from purely offline training, last centimeter errors
in interacting with objects, and mis-fitting to the task context rather than to
the actual task.  This motivates the design of our modular approach where we a)
separate out task inference (what to do) from task execution (how to do it),
and b) develop data augmentation and generation techniques to mitigate
mis-fitting.  The former allows us to leverage hand-crafted motor primitives
for task execution which side-steps the DAgger problem and last centimeter
errors, while the latter gets the model to focus on the task rather than the
task context. Our model gets 100\% and 48\% success rates on two recent
benchmarks, improving upon the current state-of-the-art by absolute 90\% and 20\%
respectively. 
\footnote{Project website with additional details: \url{https://matthewchang.github.io/awda_site/}}

\end{abstract}

\section{Introduction}
Consider a single video, demonstrating the task depicted in
\figref{problem}~(left).  Given just this input, as humans we can reliably
execute the demonstrated task in the novel situation shown in
\figref{problem}~(right). This is in spite of the differences in the task instance (location of relevant objects are different from where they were
in the demonstration), embodiment (\eg robot hand in demonstration {\it vs.}
human hand), and the large ambiguity in what precisely the task was
(was it to move the hand through those locations or was it to move the object).
In this
paper, we seek to imbue robotic agents with a similar capability: given a {\it single
visual demonstration} of a {\it novel task}, the robot should execute the
demonstrated task on a {\it novel instance} of the task. We refer to this
problem as {\it one-shot visual imitation}.

While humans are adept at this form of one-shot visual imitation, machine
performance in this setting lacks considerably. For instance, the recent method
from Dasari \etal~\cite{dasari2020transformers} obtains a 10\% success rate on
a harder version of their pick-and-place task-set, and 28\% on a one shot
visual imitation benchmark constructed using \metaworld~\cite{yu2020meta}. 
In this paper, we investigate 
what causes recent methods to underperform
and develop algorithms to bridge this performance gap.

We start by analyzing the behavior of current methods. Current works on this
problem~\cite{dasari2020transformers, finn2017one, mandi2021towards} cast it as
a conditional policy learning problem (\ie predict the {\it next action}
conditioned on the demonstration and the execution so far) using
meta-learning~\cite{finn2017one} or expressive neural network
models~\cite{dasari2020transformers, mandi2021towards}. Models are trained on
{\it offline} datasets of video demonstrations paired with expert executions.
This immediately reveals two issues that hinder the performance of these past
works. Purely offline and non-interactive training causes the learned policies
to suffer from a form of distribution shift known as the {\it DAgger problem},
(going off-distribution due to compounding errors while imitating
long-horizon behaviors) and near misses while executing fine motor control, or {\it last centimeter errors} (prior work has shown that learning generalizable policies for fine-motor control requires specialized architecture or thousands of online samples). 

\setlength{\intextsep}{0pt}%
\setlength{\columnsep}{4pt}%
\begin{figure}[t]
    \centering
    \includegraphics[width=0.48\textwidth]{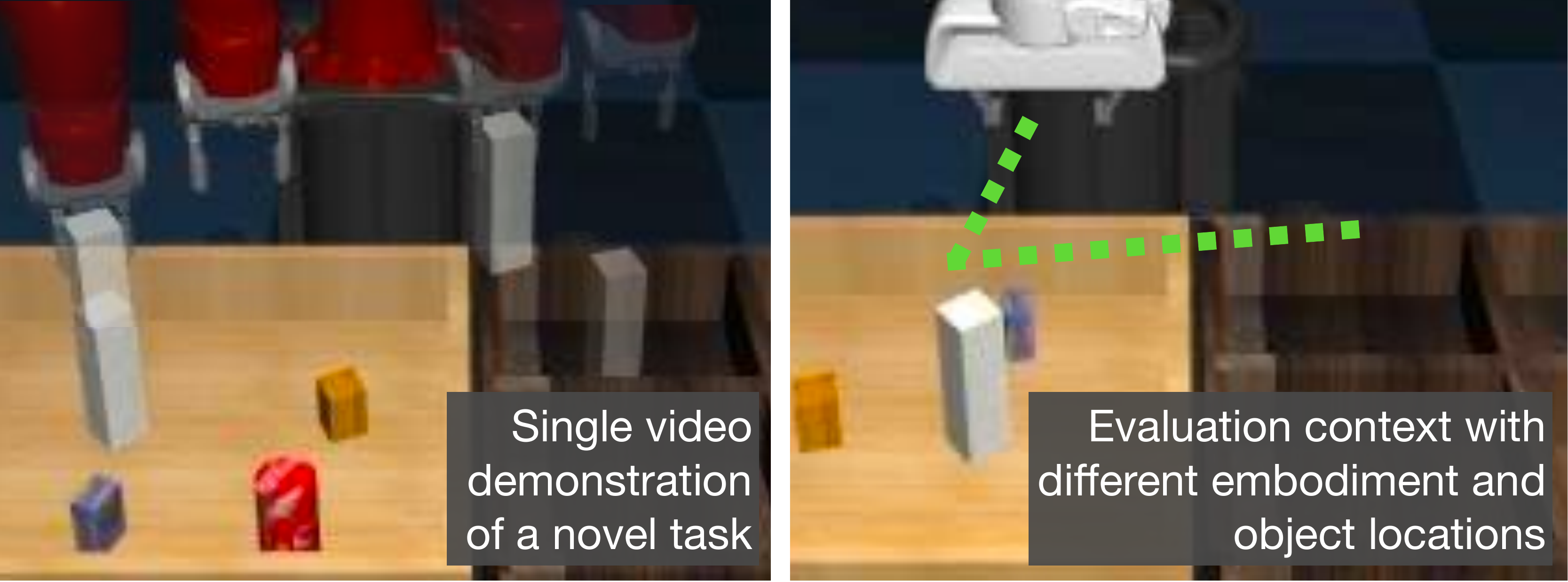}
    \caption{The One-shot Visual Imitation Problem.}
    \vspace{-7mm}
    \figlabel{problem}
\end{figure}

When we
try to extend current methods to more diverse collection of tasks, a third,
more subtle {\it mis-fitting} issue comes to light.
As tasks are often contextual (\ie one only interacts with a given 
object in a limited number of ways), 
current models tend to make predictions based on the objects in the scene 
rather than the motion depicted in the demonstration.
This causes them to generalize poorly to novel tasks. 

These insights motivate the design of our method. To circumvent the first
two problems, we employ a hierarchical and modular approach that separates out
the task execution ({\it how} to do it) from task inference ({\it what} to do).
This separation enables us to use robust and high-performing, hand-crafted motor
primitives for task execution, while the use of learning for task inference 
allows the system to interpret the intent depicted in the provided demonstration,
and synthesize a solution for the novel instance at test-time.
More concretely, given the video demonstration and just a single image of the
current scene, our learned model predicts a sequence of \textit{attributed waypoints} that outline a trajectory to achieve the task. These \textit{attributed waypoints} represent the 3D motion of the arm, along with additional attributes of the robot's state (such as ``an object is in the gripper'') at those waypoints. 
The predicted attributed waypoints are achieved using motor primitives 
(based on kinematic planning  or classical grasping primitives using depth 
images from hand-in-eye cameras). %

While this seemingly simple model works well for
pick-and-place tasks (achieving 100\% success rate on the task-set from Dasari \etal~\cite{dasari2020transformers}), it still underperforms on the diverse tasks in \metaworld~\cite{yu2020meta}, due to the mis-fitting
issue described above. 
To mitigate this, we propose novel demonstration augmentation schemes that
generate training samples to break the correlation between tasks and their contexts. 

{
We evaluate our proposed method on 4 benchmarks, representing a wide range of diverse tasks in simulation, and evaluation on real-world data~\cite{jang2022bc}. In comparisons against 3 past methods (DAML~\cite{yu2018one}, \tosil~\cite{dasari2020transformers}, and MOSAIC~\cite{mandi2021towards}) our model achieves strong results, surpassing all baselines. Notably, our method makes large improvements on two benchmarks, reaching success rates of 100\% and 48\%. These represent absolute improvements of 90\% and 20\% respectively, over the current state-of-the-art.
}

\section{Related Work} 
\noindent\textbf{Learning from demonstrations} in robotics has taken many different forms over the
years. 
One such setting is behavior cloning (BC), in which one is given
many paired sensor and action trajectories for a single task, and
the goal is to obtain a policy for this task ~\cite{pomerleau1988alvinn}.
Purely training on 
offline datasets of expert behavior
is known to suffer
from compounding errors at execution time, motivating
improvements like DAgger~\cite{ross2011reduction}. Recent approaches to this problem have considered BC using only one trajectory with action labels~\cite{duan2017one,finn2017model}, and using meta-learning to adapt to novel tasks at test time~\cite{finn2017one}.
\\\textbf{One-shot visual imitation} takes this problem a step further, where only a video
(\ie with no associated actions) of the expert's execution is available
~\cite{finn2017one, yu2018one, pathak2018zero}. Researchers have explored many
different variants, with demonstrations differing in embodiment~\cite{finn2017model, yu2018one}, viewpoint~\cite{sharma2019third}, or using natural language~\cite{jang2022bc, ahn2022can}.  Researchers have also
pursued many different solutions: conditioning on task
embeddings~\cite{james2018task, bonardi2020learning}, using meta-learning
\cite{finn2017model,yu2018one},
predicting sub-goals~\cite{pathak2018zero, sharma2019third}, using expressive
transformer architectures~\cite{dasari2020transformers, mandi2021towards}, 
and contrastive training of visual features~\cite{mandi2021towards}. 
Huang \etal~\cite{huang2019continuous} and Sharma \etal~\cite{sharma2019third}
follow a hierarchical design similar to ours for one-shot visual imitation. However, our formulation can
deal with tasks involving arbitrary objects and motions, unlike the method from
\cite{huang2019continuous} which only operates within a pre-defined set of
discrete symbols and motions.
\cite{sharma2019third} synthesizes images to use as sub-goals, while our use of
generalized waypoints sidesteps the need for image generation, which can be
challenging for novel objects in complex environments. 

We distinguish from another related line of work on inverse reinforcement
learning (IRL)~\cite{smith2019avid, zakka2022xirl, jin2020geometric,
xiong2021learning, ho2016generative,bahl2022human}. IRL assumes interactive access to the
underlying environment to learn policies, which our setting does not.
\\{\bf Hierarchical policies} have been found to useful in many settings, indoor navigation~\cite{bansal2019combining, chaplot2020learning, meng2019neural}, self-driving~\cite{muller2018driving}, drone  control~\cite{kaufmann2018deep, kaufmann2019beauty}, and manipulation~\cite{nair2018visual,finn2017deep,chitnis2020efficient, dalal2021accelerating, kaelbling2010hierarchical, pirk2020modeling} among many others.
In reinforcement learning settings, motion primitives have been incorporated as additional actions to speed up learning~\cite{nasiriany2021augmenting, dalal2021accelerating,
chitnis2020efficient}. 
Instead, our work develops techniques to use motor primitives for one-shot visual
imitation. While we found hand-designed primitives effective in our work,
our method is agnostic to the form of motor primitives, and could benefit from the
many recent works on discovering motor primitives from diverse
trajectories~\cite{kumar2019learning, shankar2019discovering,
shankar2020learning, kipf2019compile, gupta2019relay}.
\\\textbf{Data augmentation} techniques have been found to be effective in
improving the generalization of learned models~\cite{chen2020simple}.
They have also been effective at improving generalization in robot
learning, \eg when learning policies via RL~\cite{srinivas2020curl}, or for
pre-training representations~\cite{laskin2020reinforcement, mandi2021towards}.
Our work also employs data augmentation techniques, inspired by
mixup~\cite{zhang2017mixup, yun2019cutmix, berthelot2019mixmatch} for improving
generalization.  However, the specific form of overfitting in our one-shot
visual imitation problem motivates the need for {\it asymmetrical} mixing of
samples to decorrelate tasks from task contexts.

\section{Diagnosing Errors Made by Current One-shot Visual Imitation Methods}
\seclabel{analysis}
The problem we are interested in is that of one-shot visual imitation. At test
time, our agent will be given a video demonstration of a task not seen during
training, and must perform the depicted task with no additional experience in
the environment. Note that, the environment configuration may be different from
that depicted in the example video, but the overall semantic task will be the same.

Samples in one-shot visual imitation learning datasets consist of: a) the
video demonstration $\mathbf{v}$; and b) a robotic trajectory, $\{(o_1, s_1,
a_1), \ldots\}$ that conducts the same task in a potentially different
situation. 
These video demonstrations are not the same trajectory as the robotic trajectory and may differ in
embodiment, or solution method, but they must be solving the same high-level task. In
fact, the pairing of video demonstrations to robotic trajectories is what defines
the notion of a task for the model being trained. 

This is commonly cast as a supervised learning problem~\cite{dasari2020transformers, mandi2021towards, finn2017one, yu2018one}, where a model is learned on demonstration-trajectory pairs, to predict the action at a given timestep, conditioned on the demonstration, and previous frames, $\pi(a_t|\mathbf{v},o_{1:t},s_{1:t})$. However, this approach can lead to undesirable behaviors on novel tasks. In this section, we characterize the failure modes of \tosil, the transformer-based per-time-step action
prediction model from Dasari \etal~\cite{dasari2020transformers} as a
representative recent method. {Specifically, we highlight 3 different failure
modes that arise in this standard approach: the DAgger problem arising from
purely offline training, last centimeter errors in interacting with objects,
and mis-fitting to the task context rather than the depicted task.}
A summary of the terms and notation used in this paper can be found in \tableref{terms}. \renewcommand{\arraystretch}{1.5}
\begin{table*}[h]
\tablelabel{terms}
\centering
\setlength{\tabcolsep}{4pt}
\caption{Definitions of phrases and terms.}\tablelabel{sup_terms}
\resizebox{\textwidth}{!}{
{
\begin{tabular}{p{3cm}|p{10cm}}
\toprule
 \textbf{Term} & \textbf{Definition}\\
 \midrule
 Task & High level objective for the agent, \ie open the window, or move the white block to the first bin \\
 Novel Task & A task which does not appear in the training dataset \\
 Task Instance & A unique configuration of objects in the scene in which a task must be performed\\
 Task Context & The objects and elements in the scene in which a task is performed. One task context may admit many different tasks, i.e. the same blocks could be pushed, pulled, stacked etc.\\
  Demonstration ($\mathbf{v}$)& A video containing only RGB images of a task being successfully performed. This may differ from the test time environment in agent embodiment, and will feature a different task instance.\\
  Robot Trajectory $\{(o_1, s_1,
  a_1), \ldots\}$& A sequence of RGB images $\{o_1,\ldots, o_T\}$ along with robot states $\{s_1,\ldots, s_T\}$ (\eg joint angles, joint velocities) and commands $\{a_1,\ldots, a_T\}$ (\eg commanded torques, or end-effector destinations)\\
  Attributed Waypoint & A point in $3+k$ space indicating a point in $\mathcal{R}^3$ and the presence of up to $k$ attributes. See 
  \secref{policy} 
  \\
  AWDA & Attributed Waypoints and Demonstration Augmentation (our method)\\
  ADM & Asymmetric Demonstration Mixup, see 
  \secref{das}
  \\
  T-OSVI & Transformers for one-shot visual imitation \cite{dasari2020transformers}\\
  TS & Trajectory Synthesis: Generating additional training samples featuring free-space motion, see 
  \secref{das}
  \\
\bottomrule
\end{tabular}}
}
\end{table*}

\\{\it Experimental Setting.} 
We consider a
harder version of the 4 object and 4 bin pick-and-place task family proposed
in~\cite{dasari2020transformers} (visualized in \figref{problem}), that has been
modified to hold out tasks as opposed to task instances as originally done
in~\cite{dasari2020transformers}. That is, of the 16 possible tasks (picking
one of the four objects and placing it in one of the four bins), we use 14 for
training and hold out 2 for testing. %
In this modified setting, the success rate for
\tosil~\cite{dasari2020transformers} drops to 10\% from the 88\% reported in
their paper. In this setting, we identify two consistent
failure modes: a) failure to reliably reach the target object (about 88\% trials)
and often times (35\% of trials) reaching a non-target object due to what we
believe to be a version of the DAgger problem, and b) near misses in grasping
the object (about 10\% of all trials in which it 
reached any object).  
\\\textbf{DAgger problem.} \tosil can be viewed as a conditional policy of the form $\pi(a_t|\bfv,o_{1:t},s_{1:t})$, trained through behavior cloning on an
offline dataset of expert executions. Behavior cloning on expert data is
known to suffer from poor execution performance due to compounding
errors~\cite{ross2011reduction}. While this may explain the low reaching
success rate for the target object, it doesn't explain the relatively high rate
with which the policy reaches and attempts to grasp a non-target object.

Our belief is that this is a task execution error. 
The policy is trained
with memory of its execution over the last 6 frames, and often it relies
more on these recent execution frames, than the demonstration. 
Consequently, we find that if the agent makes small errors early during execution, subsequent behavior is more in line with the current execution, at the cost of being inconsistent with the given demonstration.

We empirically verify this
by keeping the demonstration fixed but {\it guiding} the policy at
test time towards the target object (positive guidance) or towards a distractor
non-target object (negative guidance), by taking steps with an oracle policy.
Even 1 step of guidance drastically improves the reaching rate from 12\%
to 58\% $\pm$ 2\%  for the target object, and from 27\% to 50\% $\pm$ 2\%
 for the distractor object. Note that it takes on
average 12 steps to reach the object, so 1 step of guidance is not much.
The increase in success rate for {\it both} target and distractor objects reveals
the preference of the policy towards past execution frames over the demonstration.
\begin{figure*}
    \centering
    \includegraphics[width=0.9\textwidth]{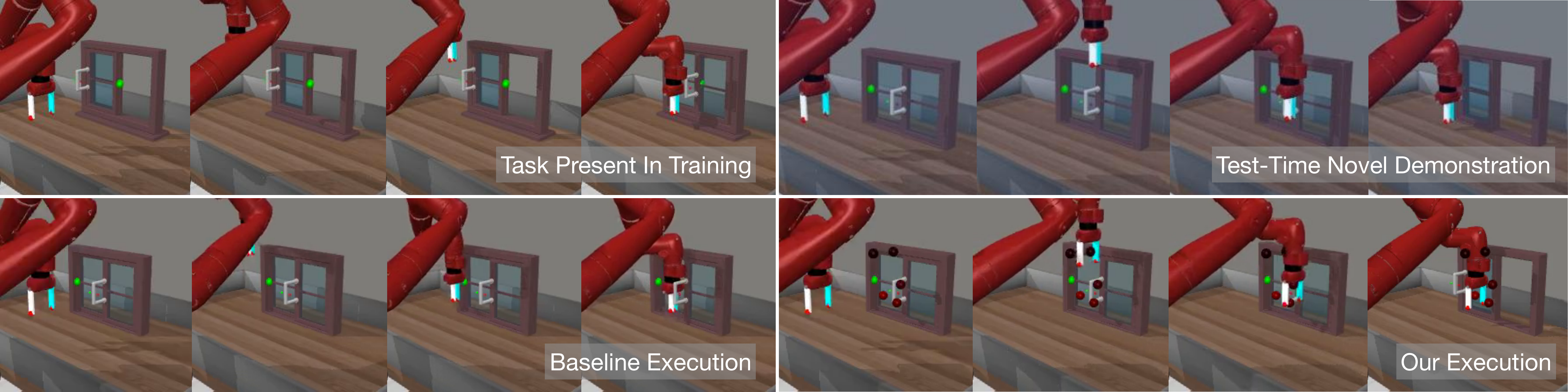}
    \caption{{\bf Example of mis-fitting in \metaworld (\secref{analysis})}: During training there is a task
    depicting the window being closed (top left). When presented with a novel
    task demonstration, opening the window (top right), action-prediction
    methods repeat the motion on the most similar training setting, trying to
    close the already closed window (bottom left). Our method 
      successfully opens the window 
    (bottom
    right). Predicted waypoints (red dots) correctly move to the right side
    of the handle and push left.}
    \vspace{-2mm}
    \figlabel{misfitting}
\end{figure*}

\textbf{Last centimeter errors in grasping.} Next, we discuss the second
substantial error mode of \tosil on this task. While 12\% of executions reach the correct object, only 10.5\% of executions successfully lift the object, meaning ~10\% of attempted grasps fail. This is because the gripper grasps near the object, but misses, or acquires an unstable grasp. 
This is not surprising as we are attempting to learn a grasping policy from 
as few as 1400 training 
samples. Past works have shown that without specialized architectures
or sensing, many thousands of trials are necessary to learn 
grasping policies that generalize~\cite{levine2018learning, pinto2016supersizing}. 
Other recent works~\cite{jang2022bc, mandi2021towards} also noted these
fine-grained errors in one-shot visual imitation. 
\\\textbf{Mis-fitting to Task Context.}
In the harder, more diverse set of 50 tasks from \metaworld~\cite{yu2020meta}, a new failure mode of one-shot visual imitation methods arises. 
{
We find that, if the novel evaluation task involves objects that are visually similar to those seen in training tasks, models trained with \tosil perform the motion from the training task, not what is seen in the demonstration (visualized in \figref{misfitting}).
}
We believe that this is because the model is predicting actions based on the task context (objects visible in the scene) as opposed to the task depicted in the demonstration. 

Tasks are contextual, \ie there are really only a few different things that one
can do with a given object (\eg opening a closed door). Besides, collecting training
data for one-shot visual imitation is challenging, as it requires paired
demonstrations and trajectories. Thus, training datasets are small and
don't showcase diverse interactions with the manipulated objects. 
Models can easily satisfy the training objective by fitting to the task context
and ignoring the motion depicted in the demonstration. This causes problems when we seek to
imitate a novel task that bears visual similarity to a training task. As depicted in \figref{misfitting}, existing methods will attempt to perform the task seen in training instead of that depicted in the demonstration.

  The analysis of these three failure modes motivates the design of our method, as detailed in the next section. 
  We address task execution errors (\ie the DAgger problem and the last centimeter problem)
  through the use of hand-crafted motor primitives, and present data augmentation strategies that break the correlation between tasks and task contexts, mitigating mis-fitting.

\begin{figure*}
    \centering
    \includegraphics[width=0.9\textwidth]{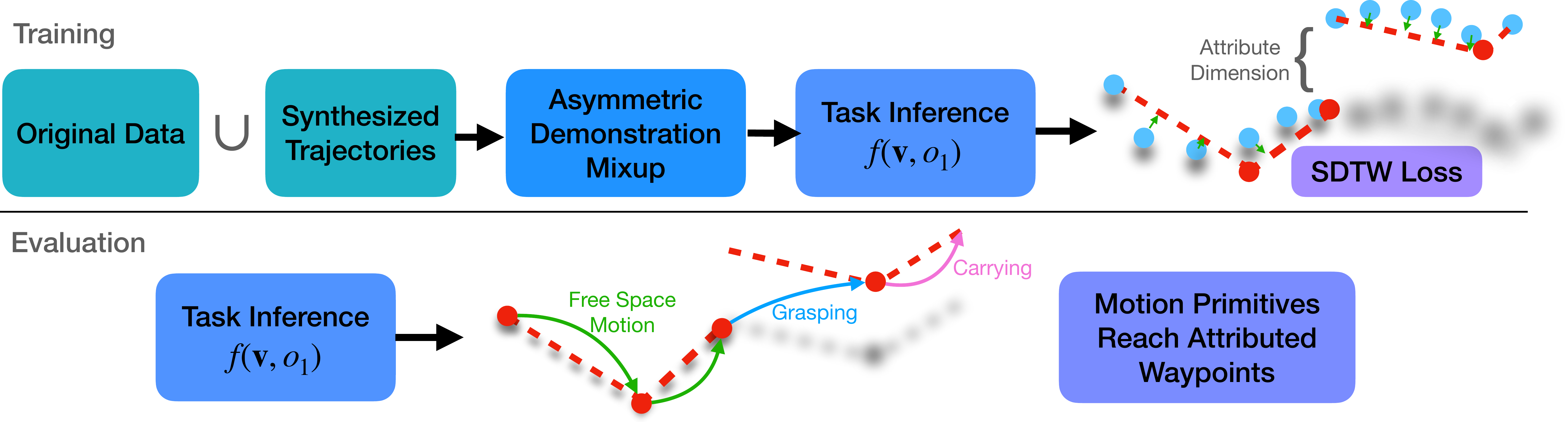}
    \caption{\textbf{Bottom:} \SM is a modular approach for one-shot 
    visual imitation that separates task inference and task execution. 
    Task inference function $f(\bfv, o_1)$ predicts a sequence of 
    {\it attributed} waypoints (red points) that are achieved using 
    hand-defined motion primitives (colored solid lines). 
    \textbf{Top Right:} $f(\bfv, o_1)$ learns to predict attributed 
    waypoints by aligning them with ground-truth attributed trajectories 
    using SDTW (\secref{policy}). \textbf{Top Left:} To
    prevent overfitting of $f(\bfv, o_1)$ to task contexts, we synthesize additional demonstrations and employ asymmetric
    demonstration miuxup (\secref{das}).}
    \vspace{-5mm}
    \figlabel{system}
\end{figure*}

\section{Visual Imitation via Attributed Waypoints and Demonstration Augmentation}
\seclabel{approach}
Our approach, \SM, is a hierarchical and modular approach that separates out task inference
and task execution. {
The task inference module takes the given demonstration video and a single image of the scene (depicting an instance of the task, different from that in the demonstration) and outputs the full execution plan, expressed as a sequence of {\it
attributed waypoints}.} Task execution happens simply by invoking the
appropriate  motor primitive to convey the robot end effector between each
consecutive pair of predicted waypoints. 
\subsection{Task Inference and Execution via Attributed Waypoints and Motor Primitives}
\seclabel{policy}
\noindent\textbf{Attributed Waypoints.} Our task inference and execution modules are
interfaced via attributed waypoints. Typical waypoints used in robotics (\eg for navigation~\cite{bansal2019combining}) only capture the 3D (or 6D)
pose of the robot end-effector. This is quite restrictive for manipulation tasks, as
purely kinematic guidance of the end-effector will not be able to interact
with objects \eg to pick them up or to exert forces on them.  We overcome this
limitation by assigning additional {\it attributes} to each waypoint, \eg is an
object in the end-effector, or is the end-effector experiencing force in a
particular direction. We consider attributed waypoints to be $3+k$
dimensional, where $k$ is the number of additional attributes associated with
each 3D waypoint. Attributed waypoints are a powerful tool for expressing solutions to kinematic tasks. For
instance, just 1 single attribute, of whether there is an object in the gripper
or not, allows us to express all 50 tasks in the \metaworld task-set as
a sequence of these 4D waypoints (3D for end-effector location and 1D for {\it ``is
there an object in the gripper''}). We will use this attribute as a
running example for explanation, but other attributes could be added.
\\\textbf{Motor Primitives.} Given a pair of attributed waypoints, our method
uses motor primitives to convey the robot between pairs of attributed
waypoints. While conveying the end-effector between 3D waypoints in space is
well understood (inverse kinematics and motion planning), moving between our
proposed attributed waypoints is more involved, as it can involve a change in
``attributes'' along the way. Thankfully, changes in attributes correspond to
well-studied basic skills in robotics literature. For instance, using the same 4D grasping example as above, going from waypoint 
$[\mathbf{p}; \text{false}]$ to $[\mathbf{q};
\text{true}]$ involves grasping an object near location $\mathbf{p}$ and taking
it to location $\mathbf{q}$; while going from $[\mathbf{q}; \text{true}]$ to
$[\mathbf{p}; \text{false}]$ corresponds to releasing the currently held object
and then going to location $\mathbf{p}$. 
In general, for $k$ attributes, this corresponds to $2^{k+1}$ motor
primitives. Our method is agnostic to the exact implementation of
motor primitives. For our experiments,
we found that hand-crafted primitives were sufficient to solve the
pick-and-place task-set from~\cite{dasari2020transformers} and all 50
\metaworld tasks~\cite{yu2020meta}. We implemented 4 hand-crafted primitives:
a) free space motion without any object in hand, b) grasping an object, c)
dropping an object, and d) free space motion with an object in hand; using
eye-in-hand depth cameras.
\\\textbf{Training the Model to Output Augmented Waypoints.}
Augmented waypoints and corresponding motor primitives let us express manipulation
tasks as a sequence of waypoints. We next describe how we train the task
inference module to predict these augmented waypoints from a given demonstration, and a single image of the novel task instance. 
Our task inference
model $f$ takes as input, a demonstration $\bfv$ and instance image $o_1$, and outputs $n$ attributed waypoints
$\{\bfw_1, \ldots, \bfw_n\}$, each waypoint being $k+3$ dimensional. 
Supervision for these waypoint predictions is derived from the trajectories in the dataset as follows.
We process the given robotic trajectory, $\{(o_1, s_1,
a_1), \ldots\}$, into 3D end-effector
locations using forward kinematics. We also assign to each time step, the appropriate attributes that the agent experiences in that frame. These attributes are not labeled by hand, but rather mined automatically from
the robot state $s_t$ and $a_t$ provided in the robotic trajectories. For example, we label that an object has been grasped in a robotic trajectory when the commanded action is to close the gripper, but the gripper jaws do not close. This gives us
end-effector's {\it attributed trajectory}: $\bft = \{\bft_1, \ldots, \bft_T\}$, $t_i \in \mathbb{R}^{k+3}$. 
We derive supervision for waypoints $\{\bfw_1, \ldots, \bfw_n\}$ by constructing a
trajectory $\hat{\bft}$ by linearly interpolating $\bfw$'s and comparing it to
the ground truth attributed trajectory $\bft$. 
We use soft dynamic
time-warping (SDTW)~\cite{cuturi2017soft} to compute the loss between the predicted
and ground truth trajectory. {
Minimizing this objective aligns the predicted trajectory with the ground truth trajectory.}  
\\\textbf{Testing.} Given a novel video $\bfv$, and task instance, as observed in
image $o_1$, we use the task inference model $f(\bfv, o_1)$ to
predict attributed waypoints. The appropriate motor primitives are used to convey
the robot from one predicted attributed waypoint to the next, until all waypoints are exhausted.
\subsection{Demonstration Augmentation for Improved Task Inference}
\seclabel{das}
We next look at tackling the {\it mis-fitting} issue highlighted in \secref{analysis}.
This mis-fitting happens because of the strong correlation between the task and
its context in the training data. 
We design two augmentation strategies that break this correlation
by generating training samples with the same context but different task motion.
\\\textbf{Asymmetric Demonstration Mixup.}
Our first strategy creates new training samples by mixing existing samples in the dataset. This is reminiscent of
mixup~\cite{zhang2017mixup} but has modifications to break the aforementioned
correlation. 
Naively mixing samples as done in original mixup wouldn't
break the correlation to aid out-of-distribution generalization.
Instead, we leverage the temporal nature of video demonstrations to \textit{asymmetrically} blend samples (depicted in \figref{mixup}).
Given a sample $(\bfv, o_1, \bft)$, we use another sample $(\tilde{\bfv},
\tilde{o}_1, \tilde{\bft})$ to produce a new decorrelated sample by: a) blending all frames in
$\bfv$ with the first frame of the video $\bf\tilde{v}$ to generate new
video $\bfv'$, b) blending $o_1$ with $\tilde{o}_1$ to generate $o_1'$, and c)
retaining $\bft$ as is. 
{
Specifically, 
\begin{equation*}
\begin{aligned}
\bfv'_t &= \alpha\bfv_t + (1-\alpha)\bf\tilde{v_0}, 
& o_1' &= \alpha o_1 + (1-\alpha)\tilde{o_1},
& \bft' &= \bft
\end{aligned}
\end{equation*}}
We use a blending ratio $\alpha \sim [0.3,
1.0]$, biased towards retaining all of
$o_1$ and $\bfv$, since the trajectory is always $\bft$.
This asymmetric blending, where one of the demonstrations is frozen in time
while the other is moving, breaks the correlation between objects present in
the scene and the task being conducted on them. $f$ can't just look at $o_1'$,
but it has to track how the hand moves through in $\bfv'$ to make correct
predictions. Including unaltered samples
in training lets us use demonstrations and observations as is at test time.
\begin{figure*}
    \centering
    \includegraphics[width=\textwidth]{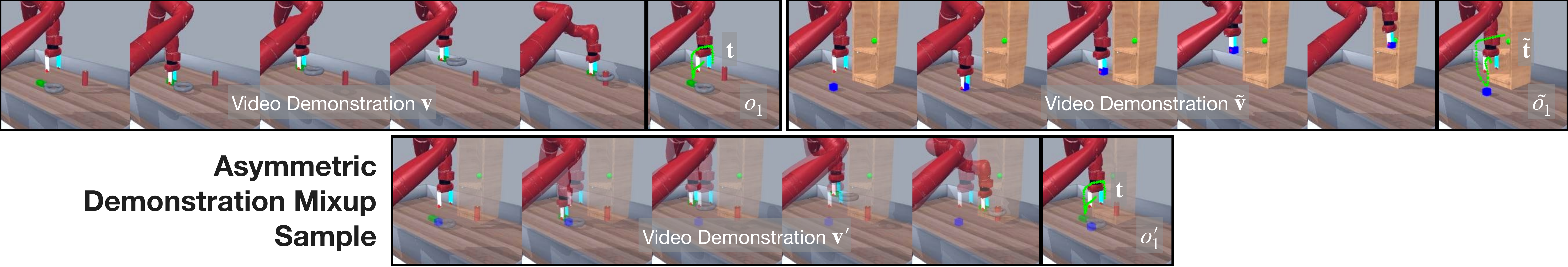}
    \caption{\textbf{Asymmetric Demonstration Mixup (ADM) augmentation:} 
    Two demonstration-trajectory pairs (row 1) are combined into one (row 2). The first frame of demonstration 2 is blended into every frame of demonstration 1. The end effector trajectory that serves as supervision for training on the combined sample is entirely from trajectory 1.}
    
    \vspace{-5mm}
    \figlabel{mixup}
\end{figure*}

\\\textbf{Additional Demonstrations via Trajectory Synthesis.}\\
Additionally, we can break the correlation between tasks and task
contexts by simply generating synthetic tasks involving free space motions for the robot, in various contexts.
We do this by sampling a small number of points (1 to 3) uniformly
at random within the agent's workspace and moving the end effector sequentially
through these points using an inverse kinematics solver. Training samples are
created by pairing each trajectory with itself, \ie $\bfv$, $o_1$ and $\bft$
all come from the same trajectory. To make correct predictions on these
trajectories, the model must attend to the motion of the arm and ignore
background elements, thus breaking the undesired correlations.  Note that
this is a simple data collection procedure that can be easily done in an
unsupervised manner. We simply add in these additional samples for training,
and find they boost performance, particularly for substantially 
out-of-distribution tasks.

We note that this procedure produces samples that are not driven by a
semantically-meaningful, object-centered task. Thus, including these samples in training could also impact the model's ability to learn a meaningful prior
over tasks. To mitigate this, we modify the final layer of the model to have two
heads. One makes predictions for original samples in the dataset, while the other makes
predictions for the synthesized trajectories. This nudges the overall network
to look at the motion of the hand while also letting the last layer learn the
necessary priors from the task driven samples in the dataset.

\section{Experiments}
\seclabel{exp}
We design and conduct experiments to demonstrate the effectiveness of our
proposed method with respect to prior work, and evaluate our various design
choices. 
\\\textbf{Tasks, Environments, and Datasets.}
We conduct experiments on 4 datasets: a) {\bf\ppe} task-set
from~\cite{dasari2020transformers} (shown in \figref{problem}) but modified to
hold 2 of 16 possible tasks as novel testing tasks; b) {\bf \metaworld}
task-set~\cite{yu2020meta} (sample observations shown in 
\figref{misfitting}) where we hold out 4 of 50 tasks as novel testing tasks;
c) {\bf MOSAIC} task-set~\cite{mandi2021towards} containing 6 tasks, evaluating performance on each task with a model trained only using demonstrations from the other tasks;
and d) {\bf BC-Z} dataset~\cite{jang2022bc} that has 17213 {\it real-world}
trajectories (sample images in \figref{bcz}) spanning 90 tasks of which we hold
out 5 for testing.

\newcommand{\tosilc}[1]{\cite{dasari2020transformers}}
\renewcommand{\arraystretch}{0.90}
\definecolor{Gray}{gray}{0.9}
\newcolumntype{g}{>{\columncolor{Gray}}c}
\begin{table*}
\centering
\setlength{\tabcolsep}{4pt}
\caption{Success rates on held-out tasks on the \ppe and \metaworld
  benchmarks 
  }\tablelabel{results}
\resizebox{\textwidth}{!}{
\begin{tabular}{lccccggggg}
\toprule
                                                        & \bf DAML& \bf \tosil & \bf Full-1 & \bf Full-2 & \bf single & \bf no AD & \bf no ADM & \bf no way & \bf only \\
                                                        &  \cite{yu2018one}                          &  \tosilc    &              &            &            & \bf head   &           &            & \bf points & \bf  waypoints \\ \midrule
Asymm. Demo Mixup (ADM)?                                &                           &                    & \cmark     & \cmark     & \cmark     & \cmark    & \xmark     & \cmark     & \xmark           \\
Additional Data (AD) Source?                            &                           &                    & TS         & BC-Z       & TS         & \xmark    & TS         & TS         & \xmark           \\
Waypoints?                                              &                           &                    & \cmark     & \cmark     & \cmark     & \cmark    & \cmark     & \xmark     & \cmark           \\
\midrule
\ppe                                                    & 0.01                      & 0.10               &\bf 1.00       &\bf 1.00       & 0.99       & 1.00      & 0.98       & 0.01       & 0.98 \\
\metaworld [easy]                                       & 0.04                      & 0.50               & 0.66       &\bf 0.73       & 0.33       & 0.74      & 0.03       & 0.33       & 0.14 \\
\metaworld [hard]                                       & 0.08                      & 0.07               &\bf 0.30       & 0.17       & 0.29       & 0.11      & 0.19       & 0.06       & 0.02 \\
\metaworld [all]                                        & 0.06                      & 0.28               &\bf 0.48       & 0.45       & 0.31       & 0.42      & 0.11       & 0.19       & 0.08 \\
\bottomrule

\end{tabular}
}
\end{table*}

\begin{figure*}[t]
    \centering
    \includegraphics[width=\textwidth]{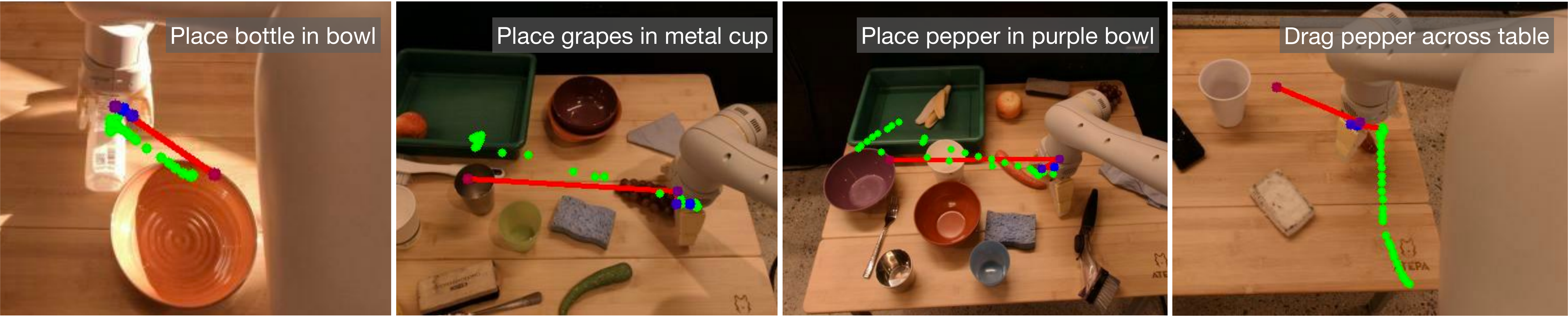}
    \caption{We visualize 2D projections of 3D waypoints in blue, and the
    interpolated trajectory in red, as predicted by our model for 4 different
    held-out tasks (noted on top right of each image) from the BC-Z
    dataset~\cite{jang2022bc}. Predicted trajectories match the ground truth trajectories (in green).
    Interestingly, for the {\it ``drag pepper across table''} task, though our
    prediction does not match the specific ground truth, it is still consistent
    with the semantics of the depicted task.}
    \vspace{-3mm}
    \figlabel{bcz}
\end{figure*}

\ppe and MOSAIC use different embodiments for demonstration and execution (Sawyer and
Panda respectively). \metaworld and BC-Z use the same embodiment. We conduct
interactive evaluation of the learned policies on \ppe, \metaworld and MOSAIC task-sets
and report success rate. For BC-Z, we do offline evaluation and report
the accuracy of predicted trajectories on a held-out validation set.
We note that as all tasks are set up in the same environment for the \ppe task
set, so it doesn't suffer from correlations between tasks and task contexts, or the mis-fitting error described in~\secref{analysis}. However,
different tasks in \metaworld and MOSAIC involve different objects making them suffer from
the mis-fitting error. 
\\\textbf{Implementation Details.} We follow ~\cite{dasari2020transformers} to
construct data for training. We collect 100 successful trajectories with each
robot using hand-defined expert controllers and arbitrarily pair them up to
construct 10K training samples per task. We train our models for 500K
iterations. Following~\cite{mandi2021towards}, we report the success rate on
held-out tasks, averaged over 5 snapshots from the end of training. Our attributed waypoints use the {\it
``is object in hand''} attribute. This leads to 4 motor primitives: free space
motion without an object, moving an object, grasping a nearby object, and releasing
the object. We implement the first two using inverse kinematics and motion
planning; the third is implemented by analyzing depth images to identify
objects and centering the gripper to grasp the nearest object; for the fourth, we just open the
gripper. 

The neural network design uses the same feature extraction process as \tosil \cite{dasari2020transformers}. We extract image features using ResNet-18, which remain spatial and have a sinusoidal positional encoding added, before being processed by a transformer module. Finally, the temporally processed features are projected down into waypoints by two separate heads, one to predict waypoints for task-driven trajectories, and one for trajectories synthesized as described in \secref{das}.
\begin{table}
\centering
\setlength{\tabcolsep}{4pt}
\caption{Success rates on held-out tasks from MOSAIC~\cite{mandi2021towards}
  task-set  }
  \tablelabel{mosaic}
\begin{tabular}{lcccccc|c}
\toprule
 \bf Task & Door   & Drawer  & Button & Blocks & B.B. & Nut A. & All   \\
\midrule
  \bf MOSAIC~\cite{mandi2021towards}      & 0.05 & 0.15  & \bf 0.05       & $0$         & $0$        & $0$ &0.04            \\
  \bf Full-1       & \bf 0.10 & \bf 0.29  & 0.01       & $0$         & $0$        & \bf 0.02 & \bf 0.07\\

\bottomrule
\end{tabular}
\vspace{-7mm}
\end{table}

\\\textbf{Results.}
We report results on the \ppe and \metaworld task sets in \tableref{results}.
We break down results on \metaworld into two splits based on the similarity of the
novel task to tasks seen in training: \metaworld~[easy] ({\it Button-Press-V2},
{\it Pick-Place-Wall-V2}, which differ from training tasks {\it Buttom-Press-Wall-V2} and {\it Pick-Place-V2} only due to presence/absence of distractor), and
\metaworld~[hard] ({\it Window-Open-V2}, {\it Door-Unlock-V2}, require categorically different solutions than training tasks). The results on the MOSAIC benchmarks are presented in \tableref{mosaic}. Each column reports the performance on the indicated held-out task. The models for MOSAIC experiments are trained on all MOSAIC tasks except the held-out task. The column {\it All} reports the mean performance across all held-out task experiments. 
We summarize our key takeaways below.\\
\textbullet~{\bf \SM outperforms prior work by a
large margin.} Our full system (denoted {\it Full-1}) completely solves the \ppe
task (improving upon the 10\% obtained by \tosil, 1\% by DAML), and obtains 48\% for
\metaworld~[all] \VS 28\% for \tosil, 6\% for DAML, while quadrupling the performance on the
hard tasks 30\% \VS 7\% for \tosil. Performance gains are maintained even if we
omit synthesized trajectory data altogether (denoted {\it no AD}), or when
using data from other datasets instead (denoted {\it Full-2}). On the MOSAIC tasks (\tableref{mosaic}), we match or outperform the current state-of-the-art for this benchmark~\cite{mandi2021towards} on all tasks except for one, yielding superior overall performance (7\% \VS 4\%).\\
\textbullet~{\bf Attributed waypoints with motor primitives eliminate all
errors on \ppe.} Our models without asymmetric demo mixup ({\it no
ADM}), or without additional data ({\it no AD}), or without both ({\it only waypoints}) obtain close to perfect performance on the \ppe task.  This
demonstrates the effectiveness of our proposed modular policy architecture.  It
also boosts performance on \metaworld~[all] by an absolute 29\% from 19\%
with {\it no waypoints} \VS 48\% with ({\it Full-1}).
\\
\textbullet~{\bf Additional data via trajectory synthesis or from other
datasets helps improve generalization}.
Using additional data via trajectory synthesis ({\it Full-1}) or from
other datasets ({\it Full-2}) improves upon not using any additional data
(denoted {\it no AD}), particularly for the \metaworld~[hard] tasks that
require entirely novel motion at test time (11\% for {\it no AD} \VS 30\% for
{\it Full-1} and 17\% for {\it Full-2}). Furthermore, fitting this additional
data through another head is crucial for maintaining performance on
\metaworld~[easy] tasks, which bear more similarity to training tasks: 66\% for
the two headed model {\it Full-1} \VS 33\% for the {\it single head} model.
\\
\textbullet~{\bf Asymmetric demonstration mixup improves performance} beyond the standard image augmentations (random flip, crop, translation, color jitter, \etc) that are
already in use for \tosil, DAML and all our models ({\it Full-1} \VS~{\it no
ADM}). \\
\textbullet~\textbf{\SM gets good performance on real data from robots.} On the
BC-z dataset~\cite{jang2022bc}, our method is able to predict the final
interaction point (grasp, release, or reach point) to within 10~cm for
$63\%\pm4\%$ samples of \textit{held-out} tasks.  This clearly identifies what
objects need to be interacted with. We expect appropriately designed motor
primitives on physical robots will be able to successfully execute some of
these tasks. \figref{bcz} shows some sample predictions.

\section{Conclusion and Future Work}
In this paper we analyzed the major failure modes of state-of-the-art action prediction methods for one-shot visual imitation. We find that they suffer from the DAgger problem, last centimeter errors, and mis-fitting to task contexts. Our proposed method, utilizing attributed waypoints and demonstration augmentation, is able to significantly boost success rates on existing benchmarks, even completely solving one.

{
As is, our system is limited to kinematic tasks, but could be expanded to reasoning about forces, given the proper motion primitives. While our motion primitives are closed-loop and account for slight changes in object locations, the high level plan cannot adjust to large changes in the scene after initial waypoint predictions. We leave this to future work.
}

\section*{ACKNOWLEDGMENT}
This material is based upon work supported by NSF (IIS-2007035), DARPA (Machine Common Sense program), an Amazon Research Award, an NVidia Academic Hardware Grant, and the NCSA Delta System (supported by NSF OCI 2005572 and the State of Illinois)

\bibliographystyle{IEEEtran}
\bibliography{biblioLong.bib,refs_full.bib}  %

\clearpage
\section*{APPENDIX}
\begin{figure*}[h]
    \centering
    \includegraphics[width=1\textwidth]{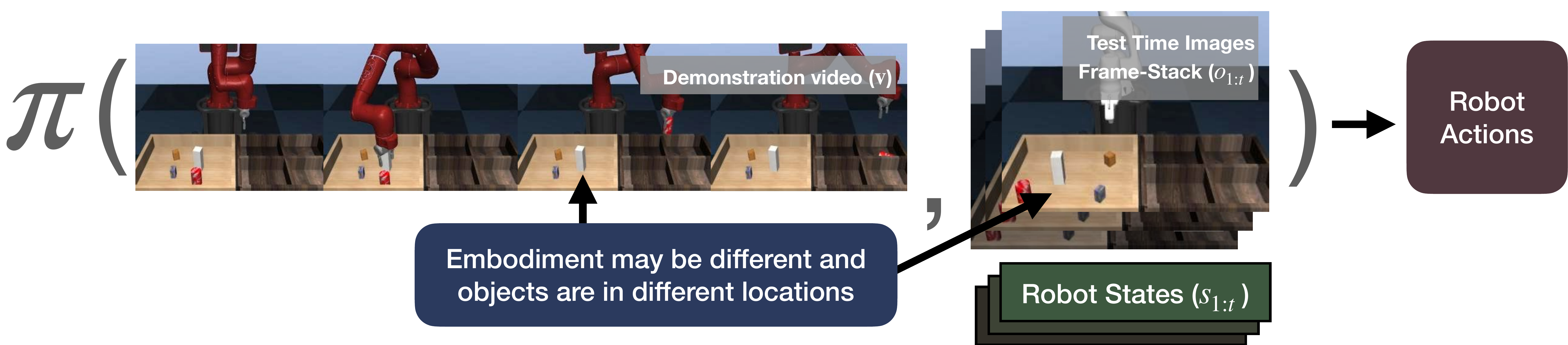}
    \caption{{\textbf{The one-shot visual imitation problem}: Given a video demonstration ($\bfv$) of an agent performing a novel task (a task not seen during training), and the sequence of images for the current execution so far ($o_{1:t}$), the policy $\pi(a_t|\bfv,o_{1:t},s_{1:t})$ must predict the next action to take to accomplish the task depicted in the video demonstration. Note that the agent embodiment and the locations of the objects may differ between the video demonstration and the test time environment.}}
    \figlabel{osil}
\end{figure*}

\section{Motor Primitives Details} \seclabel{sup_primitives} For the
experiments in this paper we utilize four motor primitives: free-space motion,
carrying, dropping, and grasping. We normalize the action spaces for all environments
to behave like the \metaworld environments. They have 4 degrees of
freedom, 3-DOF delta end-effector position control, and 1-DOF for desired
gripper closure. We implement 3-DOF delta end-effector position control by 
driving the joints to a configuration that achieves the desired end-effector
position as obtained through inverse kinematics. 

\textbf{Free-space Motion: } Given the current end-effector position and a
destination, for each time step we move the end-effector directly towards the
destination point in a straight line, using the abstracted delta end-effector
position control space defined above.

\textbf{Carrying:} Carrying is implemented using the above free-space motor
primitive, just keeping the gripper closed throughout the movement.

\textbf{Dropping: } Dropping is implemented by simply opening the gripper in place.

\textbf{Grasping: }
We utilize depth images from an eye-in-hand camera for the grasping primitive.
At a high level, the grasping primitive attempts to localize the nearest object
to the gripper, position the gripper above that object, then drop down to
perform the grasp.  We first outline the steps of the motor primitive, and then
describe the object localization procedure.
\begin{enumerate}
  \item Determine the target object position using depth images (described below).
  \item Move the gripper to a point 15cm above the estimated object position,
  recomputing the object position at each time-step.
  \item After reaching a point above the object, move the gripper down to the
  estimated object $z$ position offset by some small value (3cm for the panda
  gripper, but this should be adjusted based on the agent end effector).
  \item Close the gripper.
  \item Lift 15cm.
\end{enumerate}

Next, we describe the object localization procedure to recover the target object
position (step 1 above):
\begin{enumerate}
  \item Mask out background values, defined as having a depth greater than 1
  meter.
  \item Determine the ground plane distance by taking the median of the
  remaining depth values.
  \item Produce a mask for potential objects, taking pixels 1cm or more above
  the floor plane.
  \item Compute connected components of the potential object mask to get
  individual object proposals.
  \item Select the connected component with the centroid closest to the center
  of the frame as the closest object (target object for grasping).
  \item Use the camera intrinsics to project the depth point of the centroid
  back into 3D world coordinates. This is the estimated object position.
\end{enumerate} 

\subsection{Mining Attributes}
\seclabel{mining}
{
Whether or not the attribute of \textit{``is there an object in the gripper''}
is present at every timestep $t$, is mined directly from the sequence of
joint angles and commanded actions $\{(s_1,a_1)\ldots\}$. We determine that an
object is held when the commanded action is to close the gripper, but the
gripper jaws do not close. At each timestep, we annotate if it potentially
contains a grasp with a variable $g_t$.  \[g_t = c_t \wedge (s^g_{t+d} -
s^g_{t} > \delta)\] Where $c_t$ is a boolean indicating if the gripper was
commanded to close on frame $t$, and $s^g_t$ is the component of the robot state at
time $t$ that corresponds to the distance between the gripper jaws. $d$ and
$\delta$ are tuned based on the timescale of the data, and dimensions of the
robot. For the \metaworld data, we use $d = 10$ and $\delta = -0.05$, for the
\tosil data we use $d = 1$ and $\delta = -0.05$. For BC-Z, we use $d = 1$,
$\delta = -0.07$. The final sequence of potential grasp frames $g_0,\ldots,g_T$
is smoothed with a convolutional filter to give the final attribute
annotations.
}

In addition to the grasping primitive used in our experiments, we have
validated that this approach can work well for other attributes. We are able to
detect a ``pressing'' attribute (\ie button presses) using a similar
methodology as for detecting object grasps. We identify frames in which the
commanded action is to move the end-effector in a certain direction, but
the end-effector does not move, or accelerate in a direction consistent with the
commanded action. This means there must be some object obstructing the
end-effector, \ie the end-effector is pressing into something. We visualize the
results of this technique in \figref{pressing}. We show the first pressing
frame detected in trajectories from four different Meta-World tasks which
require pressing ({\tt coffee-button-v2}, {\tt button-press-v2}, {\tt
handle-press-v2}, and {\tt button-press-topdown-v2}). We find that this form of
automatic mining is able to accurately identify frames in which the agent is
performing button presses or other pressing-like actions.
\begin{figure*}
    \centering
    \includegraphics[width=\textwidth]{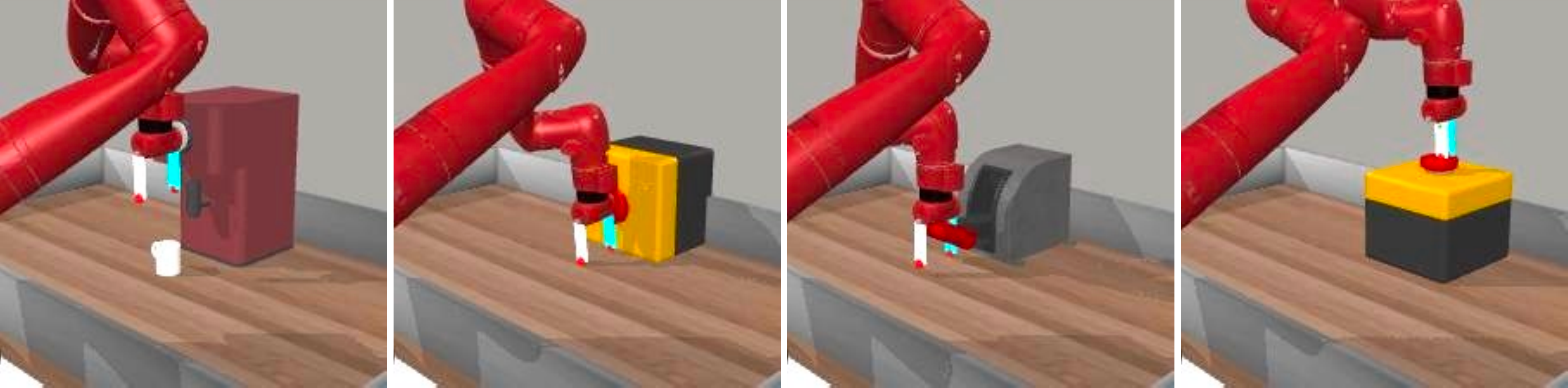}

    \caption{The first frames in which a ``pressing'' attribute is detected in
    four trajectories from four different task in the Meta-World task-set, from
    left to right: {\tt coffee-button-v2}, {\tt button-press-v2}, {\tt handle-press-v2}, and {\tt button-press-topdown-v2} } \vspace{-5mm}
    \figlabel{pressing}
\end{figure*}

\section{Implementation Details}
\seclabel{sup_arch}
\subsection{Architecture}
The architecture has three modules as shown in \figref{arch}: a convolutional
image feature extractor, a transformer with self-attention layers for temporal
processing, and a multi-headed MLP for waypoint prediction.  These first two
modules (feature extractor and transformer) are identical to the \tosil
architecture. Our architecture differs only in the waypoint prediction heads.
We notate the observation of the novel task instance as $o_1$, as only the first frame
is used in our method. For the baselines, we use a frame-stack of 6 images
$\{\ldots,o_{t-1},o_t\}$, following \tosil.

\textbf{Image feature extraction}:
Visual features are extracted by a pre-trained ResNet-18~\cite{he2016deep}. We
share weights between the feature extractor for $\mathbf{v}$, and ${o_1}$,
frames. We remove the last two layers of the ResNet and use the last spatial
features.

\textbf{Self-Attention Layers}:
We pass these spatial features into the self-attention module. We stack spatial
features together along the time dimension, with ${o_1}$
(and any frame-stacking for non-waypoint baselines) placed after $\mathbf{v}$.
We add sinusoidal positional encodings to the features, treating time
and space as a single dimension. We then pass these spatial features with positional
encodings into a multi-headed self-attention module. We create key, query, and value
tensors by applying 3D convolutions with kernel size 1.
We then apply dropout, a residual connection, and batch normalization after each
self-attention layer.  Following \tosil, all methods use 4 attention heads and
2 layers of self-attention. Finally, after the attention layers, we apply a spatial
softmax, followed by a two-layer MLP. This projects the features from
the final time-step, corresponding to the ${o_1}$, into a fixed
length vector.

\begin{figure*}
    \centering
    \includegraphics[width=0.8\textwidth]{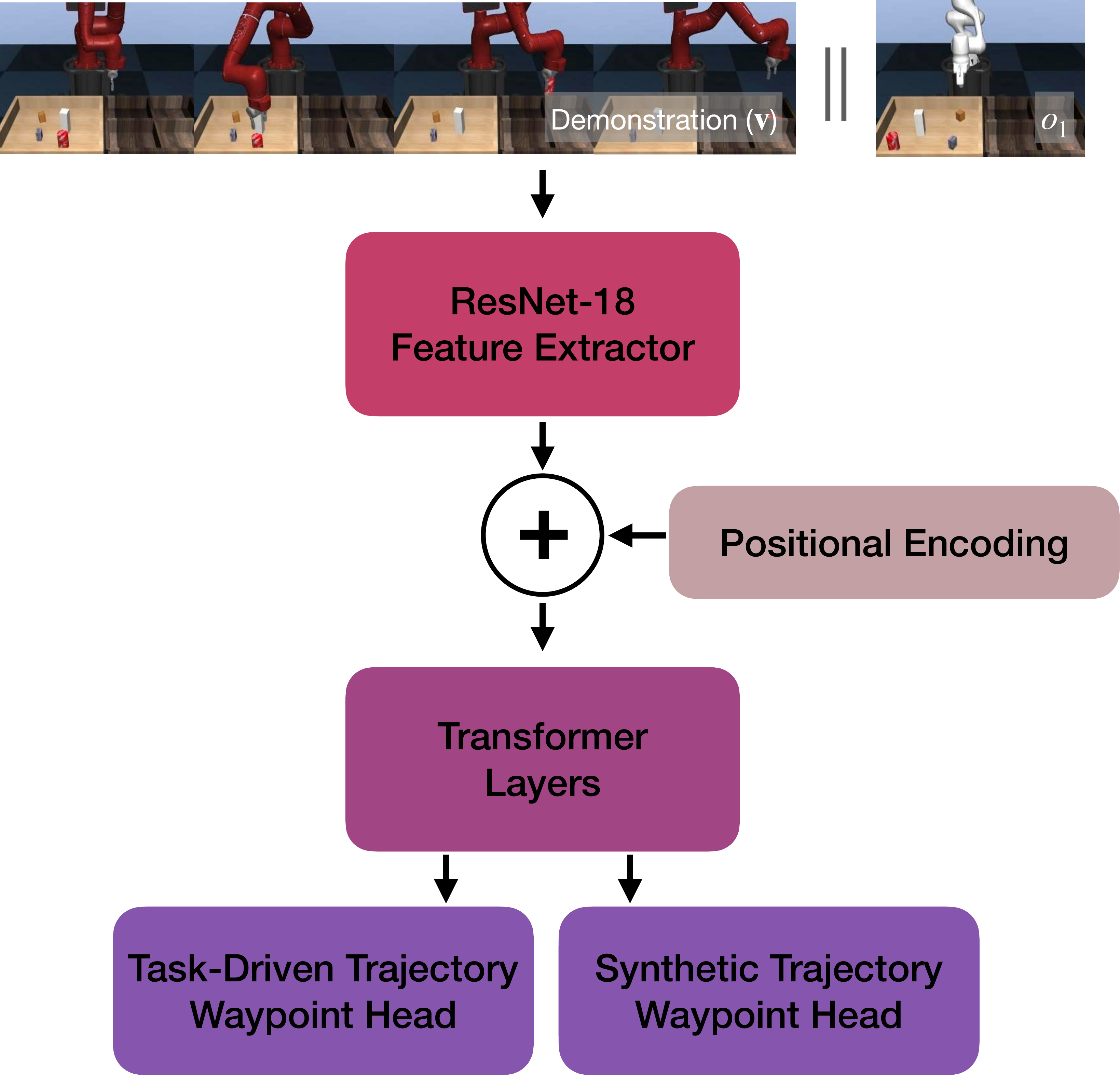}
    \caption{Outline of our model architecture. Following \tosil [1], we extract features using ResNet-18. These features remain spatial, and have a sinusoidal positional encoding added, before being processed by a transformer module. Finally, the temporally processed features are projected down into waypoints by two separate heads, one to predict waypoints for task-driven trajectories, and one for trajectories synthesized as described in 
    \secref{das}
    }
    \vspace{-5mm}
    \figlabel{arch}
\end{figure*}

\textbf{Waypoint Prediction}:
We pass the final features (fixed-length vector corresponding to $o_1$ from
above) through a multi-headed 2-layer MLP, \ie, there is a separate layer to
transform the features into waypoints for task-driven samples and synthesized
samples. 
We select predictions from one head or the other based on the dataset of origin
(task-driven or synthesized) of each trajectory in the batch.

\subsection{Loss Computation}
We linearly interpolate between the waypoints to produce a set of points
amendable for use with Soft Dynamic Time Warping (SDTW) [39] with a fixed temperature
parameter of $0.001$. All experiments downstream are conducted using 5
waypoints. However, at training time we predict a total of 5
trajectories consisting of 1, 2, 3, 4, and 5 waypoints respectively, for a total of
15 waypoints. The SDTW loss against ground truth is computed independently for
each of these 5 trajectories. The final loss is the mean across the computed
SDTW distance for all 5 trajectories.

\subsection{Inference Speed}
{
When running our model on a single modern GPU (Nvidia A40), our model is able to make a forward pass through the network to predict waypoints in 7 ms. After waypoints are predicted, using the motor primitives to command the agent requires less than 2ms per environment step.
}
\begin{table}[h]
\centering
\setlength{\tabcolsep}{4pt}
\caption{Hyper-parameters for training and evaluation}
  \tablelabel{sup_hyperparams}
\begin{tabular}{ll}
\toprule
\textbf{Hyperparameter} & \textbf{Value}\\
\midrule
Image size (\metaworld) & (224, 224)\\

Image size (\ppe) & (240, 320)\\
Learning Rate & $0.0005$ \\
Batch Size & 30 \\
Non-Local Attention Layers & 2 \\
Attention Heads per Layer & 4 \\
\# of Waypoints for Evaluation & 5 \\
Evaluation Episodes per Snapshot per Task & 20 \\
$\gamma$ (SDTW Temperature) & $0.001$ \\
\# of Demonstration ($\mathbf{v}$) frames & 10 \\
Optimizer & Adam \\
Post-Transformer Hidden Dimension & 256 \\
\bottomrule
\end{tabular}
\end{table}

\section{DAgger Problem Continued Analysis}
\seclabel{sup_dagger}
One may wonder about the role of action sampling and frame-stacking in our
analysis of the DAgger problem in 
\secref{analysis}.
We investigate this by evaluating the baseline (with frame-stack 6) by taking
the mean action and mode action. Additionally, we train a model using no
frame-stacking, relying only on the current frame for prediction. We find that
none of these modifications address the failure mode studied in 
\secref{analysis}.

When training with no frame-stack we see the same poor performance, with this
model achieving a reaching success of 0.12. We believe this is because, in any
state where the recent frames dictate which object to reach, only the current
frame is sufficient to make the same prediction, \ie, the model can just look
at which object the gripper has moved towards from the current frame. It
doesn't need the past frames for this. Additionally, we find the model with no
frame-stack achieves a significantly \textit{lower} overall success rate than
the baseline down to 0.01 from 0.1. This is because the frame-stack is useful
for performing the grasp on the object. Demonstrations pause the griper above
the object before grasping, and the model is unable to fit this behavior well
without the previous frames: it pauses above the object and never descends to
attempt a grasp. Using the mean or mode action did not resolve
the reaching issue either, achieving a reaching success of 0.11 and 0.10
respectively.

\section{Grasping Failure Details} 
Grasping analysis results (as reported in 
\secref{analysis}
, and above) are computed
over 440 evaluations on the best snapshot from the action-prediction baseline
\tosil. We consider the object to be successfully reached when the end effector
comes within 4cm of the object and no grasp has yet been attempted in the
current episode. A grasp is considered successful when the object has been
reached, and then the object is raised at least 5cm off the ground. 

As mentioned in 
\secref{analysis}
, even when the right object is reached, grasping
control in an end-to-end learned model with limited data is often
unreliable. In \figref{sup_grasp_failures} we visualize two of the typical
grasping failure modes: missing the grasp, or acquiring an unstable grasp which
drops the object.

\begin{figure*}[h]
    \centering
    \includegraphics[width=0.8\textwidth]{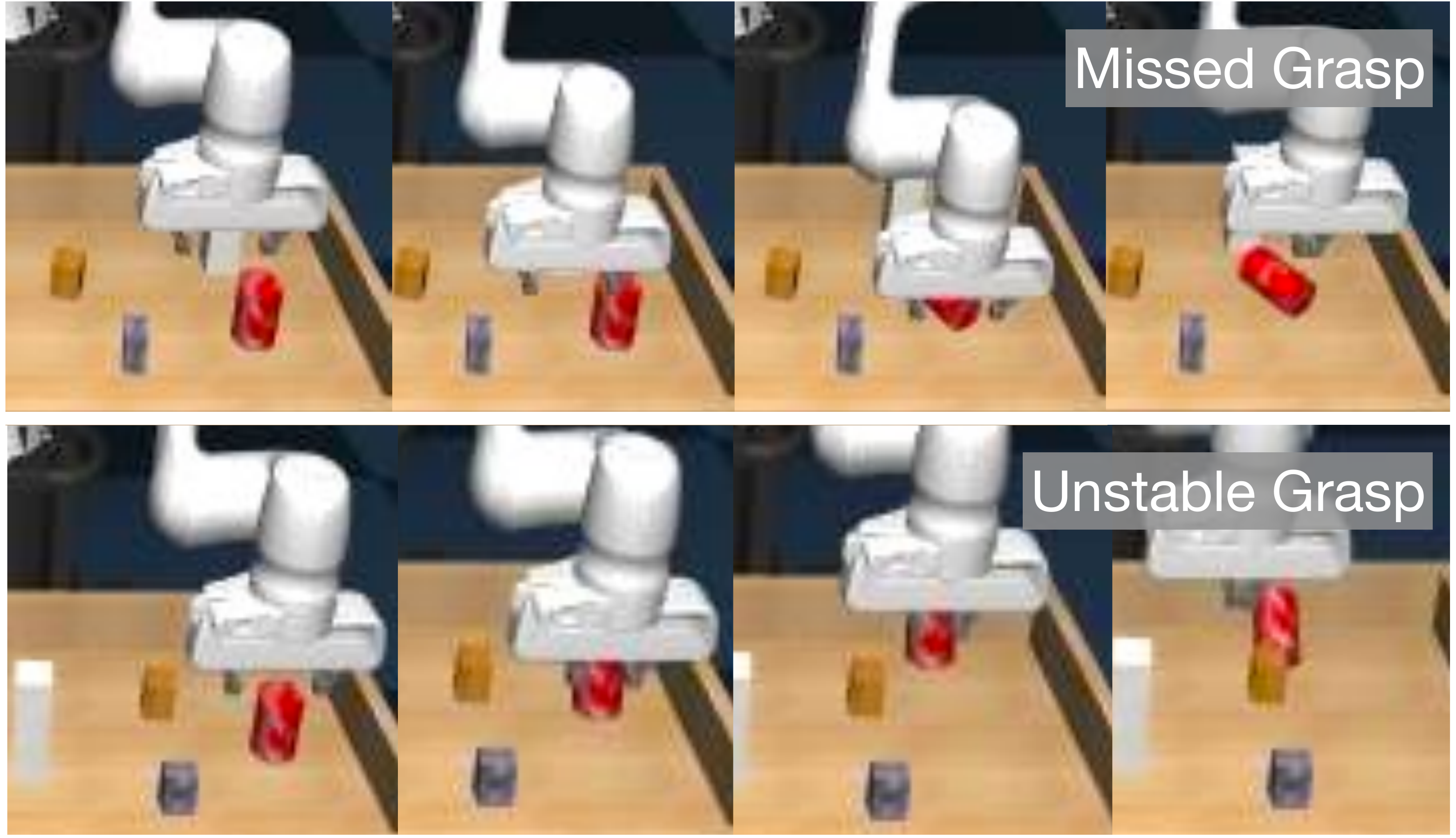}
    \caption{The two most common grasping failure modes: missing the grasp
    entirely, and acquiring an unstable grasp.}
    \figlabel{sup_grasp_failures}
\end{figure*}

\section{Experimental Details and Results} \seclabel{sup_exps} In
\tableref{sup_results} we report mean success rates with standard error over
the 10 evaluated snapshots of our method. In addition to methods reported in
the main paper, we include {\bf only waypoints}, which has neither additional
data, nor our asymmetric demonstration mixup augmentation, and {\bf
\tosil\nolinebreak +depth}, which is \tosil augmented to include the same eye-in-hand depth
images that our method uses for grasping. We do this by adding a separate
ResNet for depth image feature extraction, and perform late-fusion, combining
RGB and depth image features into one vector before transformer layers. This
performs significantly worse, likely because this adds more
capacity to over-fit, following our analysis in 
\secref{analysis}
. In
\tableref{mosaic_detail}, we report the error bars for the MOSAIC results from
\tableref{mosaic}.
We follow the same procedure as other environments,
reporting the mean and standard error of success rates over the last 10
snapshots from training. However, the results from
MOSAIC~\cite{mandi2021towards} only report standard deviation over the last 3
snapshots from training. For fairer comparison we have computed the estimated
standard error based on their reported standard deviations assuming normality of
the underlying samples.
\renewcommand{\arraystretch}{1.05}
\begin{table*}[h]
\centering
\setlength{\tabcolsep}{4pt}
\caption{We report mean success rates and standard error on held-out tasks on the \ppe and \metaworld
benchmarks.}\tablelabel{sup_results}
\resizebox{\textwidth}{!}{
\begin{tabular}{lccgccggggg}
\toprule
                                                                  & \bf DAML~\cite{yu2018one} & \bf \tosil~\tosilc & \bf \tosil    & \bf Full-1    & \bf Full-2    & \bf single    & \bf no AD     & \bf no ADM    & \bf no way          & \bf only way       \\
                                                                  &                           &                    & \bf +depth    &               &               & \bf head      &               &               & \bf points          & \bf  points \\ \midrule
  \multicolumn{2}{l}{Asymm. Demo Mixup (ADM)?}                    &                           &                    & \cmark        & \cmark        & \cmark        & \cmark        & \xmark        & \cmark        & \xmark           \\
  \multicolumn{2}{l}{Additional Data (AD) Source?}                &                           &                    & TS            & BC-z          & TS            & \xmark        & TS            & TS            & \xmark           \\
  \multicolumn{2}{l}{Waypoints?}                                  &                           &                    & \cmark        & \cmark        & \cmark        & \cmark        & \cmark        & \xmark        & \cmark           \\
\midrule
\ppe                                                              & $.01 \pm .00$             & $.10 \pm .05$      & $.00 \pm .00$ & $1.0 \pm .00$ & $1.0 \pm .00$ & $.99 \pm .00$ & $1.0 \pm .00$ & $.98 \pm .01$ & $.01 \pm .02$       & $.98 \pm .01$  \\
\metaworld [easy]                                                 & $.04 \pm .01$             & $.50 \pm .08$      & $.02 \pm .01$ & $.66 \pm .05$ & $.73 \pm .08$ & $.33 \pm .06$ & $.74 \pm .05$ & $.03 \pm .01$ & $.33 \pm .10$       & $.14 \pm .08$  \\
\metaworld [hard]                                                 & $.08 \pm .03$             & $.07 \pm .03$      & $.02 \pm .01$ & $.30 \pm .04$ & $.17 \pm .04$ & $.29 \pm .03$ & $.11 \pm .04$ & $.19 \pm .06$ & $.06 \pm .02$       & $.02 \pm .01$  \\
\metaworld [all]                                                  & $.06 \pm .01$             & $.28 \pm .06$      & $.02 \pm .01$ & $.48 \pm .05$ & $.45 \pm .08$ & $.31 \pm .03$ & $.42 \pm .08$ & $.11 \pm .04$ & $.19 \pm .06$       & $.08 \pm .04$  \\
\bottomrule
\end{tabular}
}
\end{table*}

\begin{table*}
\centering
\setlength{\tabcolsep}{4pt}
  \caption{Success rates on held-out tasks from MOSAIC~\cite{mandi2021towards}
  task-set }
  \tablelabel{mosaic_detail}
\resizebox{\textwidth}{!}{
\begin{tabular}{lcccccc|c}
\toprule

 \bf Task & Door   & Drawer  & Press Button & Stack Block & Basketball & Nut Assembly & All   \\
\midrule
  \bf MOSAIC~\cite{mandi2021towards}$^1$
             & $0.05 \pm 0.018$         & $0.15 \pm 0.038$              & \bf $\mathbf{0.05} \pm 0.018$ & $0$ & $0$ & $0$                          & $0.04 \pm 0.008$          \\
  \bf Full-1 & $\mathbf{0.10} \pm 0.02$ & \bf $\mathbf{0.29} \pm 0.033$ & ${0.01} \pm 0.005$            & $0$ & $0$ & \bf $\mathbf{0.02} \pm 0.01$ & $\mathbf{0.07} \pm0.007$\\

\bottomrule
\end{tabular}
}
\\
\vspace{2mm}
{\tiny \textsuperscript{1} The results from
MOSAIC~\cite{mandi2021towards} only report standard deviation over the last 3
snapshots from training, where we report standar error over the last 10 snapshots for a lower variance estimate. For fairer comparison we have computed the estimated
standard error based on their reported standard deviations assuming normality of the underlying samples.}
\vspace{-7mm}
\end{table*}

\section{Visualizations}
\seclabel{sup_vis}
\figref{examples} shows the waypoint predictions of our model on novel
tasks. Notice that our method mimics the overall solution scheme of the
demonstration while adapting to the different environment configuration in the
current scene. We visualize trajectories from our trajectory
synthesis method from 
\secref{das},
in \figref{traj}.
\begin{figure*}
    \centering
    \includegraphics[width=\textwidth]{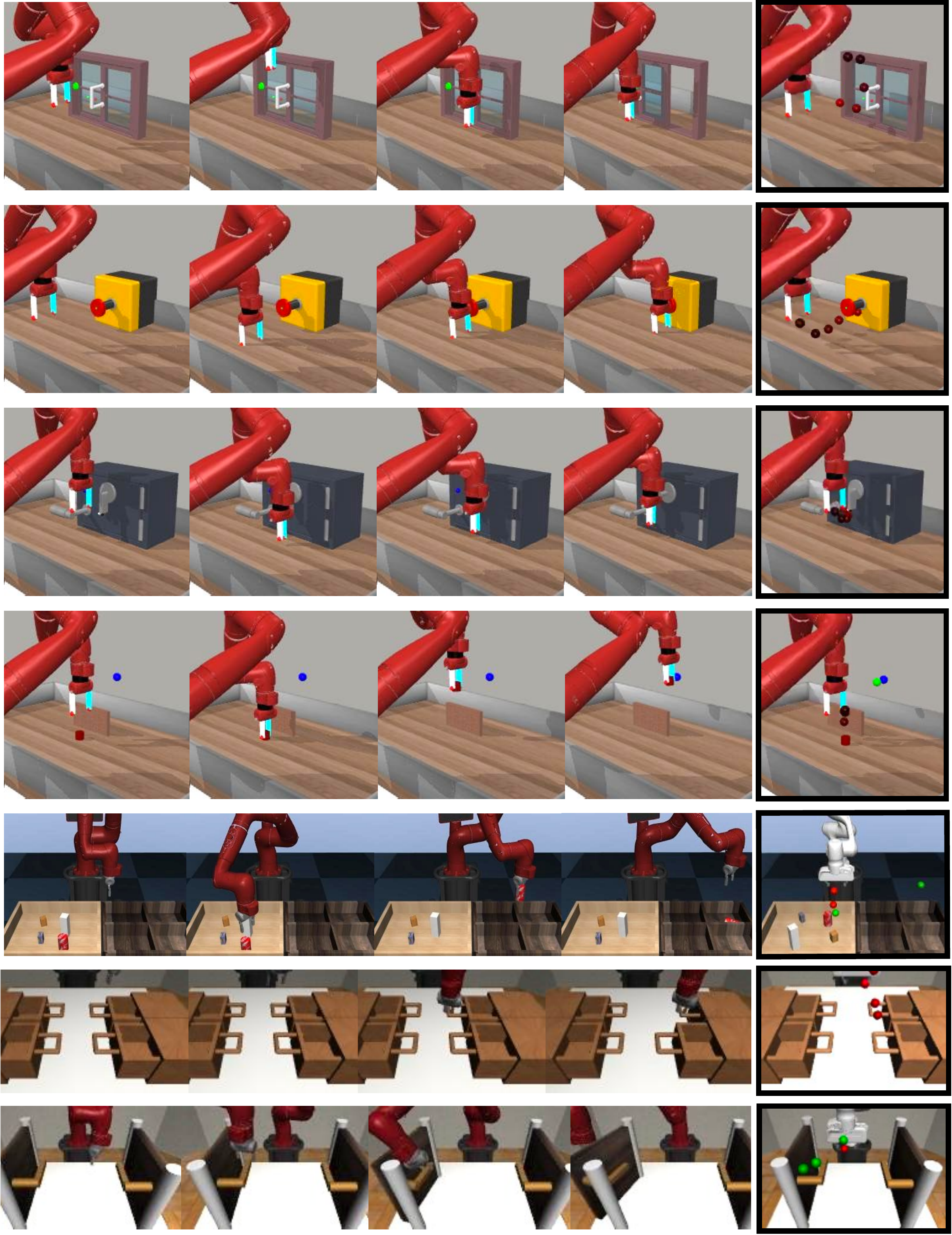}
    \caption{We visualize example waypoint predictions on held-out tasks on
    Metaworld, \ppe, and MOSAIC. Each row is a different novel task that the model has
    not seen during training. The frames on the left represent the video
    demonstration, and the frame on the right shows the waypoints predicted in
    a new setting. Red spheres represent free-space waypoints. The increasing order of
    waypoints is indicated by a brightening of the color. Green spheres
    indicate the end point of line segments attributed with the \textit{object
    grasped} attribute. This means the grasping primitive should be invoked at
    the previous waypoint if it is red, and the object should be carried to the
    green waypoint.}
    \figlabel{examples}
\end{figure*}

\begin{figure*}
    \centering
    \includegraphics[width=0.8\textwidth]{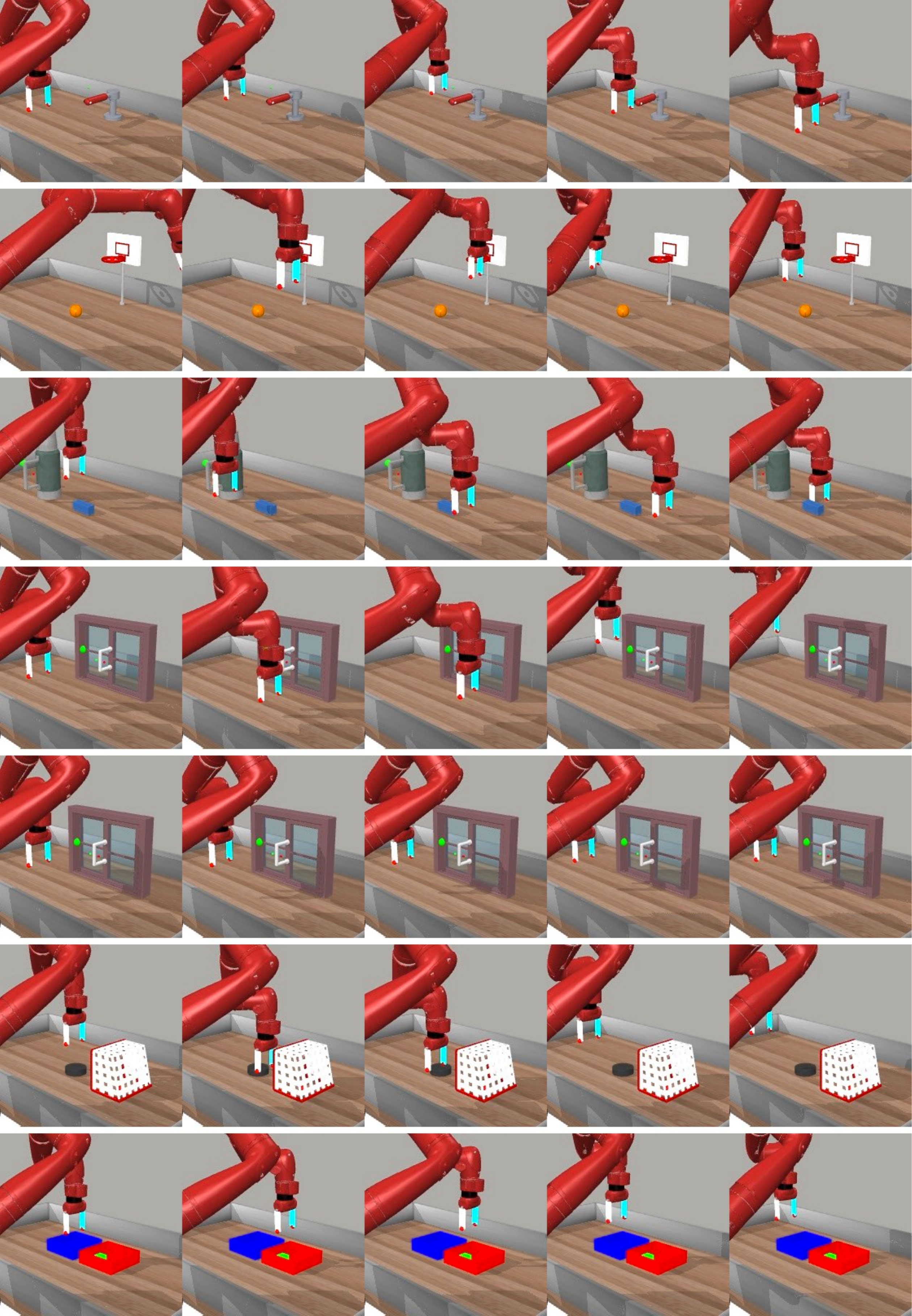}
    \caption{We visualize example trajectories generated by our proposed
    trajectory synthesis method (
    \secref{das}
    ). Trajectories are synthesized by
    driving the end-effector through randomly sampled waypoints in the
    environments used for training. Introducing this data into the training
    dataset helps decorrelate tasks from task contexts.}
    \vspace{-5mm}
    \figlabel{traj}
\end{figure*}

\end{document}